\documentclass{ws-procs11x85}
\usepackage{ws-procs-thm}   
\usepackage[numbers]{natbib}
\usepackage{longtable}
\usepackage{array}
\usepackage{multirow}  % Needed for the 'multirow' command
\usepackage{colortbl}  % For coloring table rows
\usepackage{xcolor}    % For defining colors
\usepackage{booktabs}

\begin{document}
\begin{minipage}{\textwidth}
\title{ReXamine-Global: A Framework for Uncovering Inconsistencies in Radiology Report Generation Metrics}

\newcommand{\compactauthor}[2]{#1\textsuperscript{#2}\space}

\author{%
\compactauthor{Oishi Banerjee}{1*}
\compactauthor{Agustina Saenz}{1*}
\compactauthor{Kay Wu}{1*}
\compactauthor{Warren Clements}{2,†}
\compactauthor{Adil Zia}{2,†}\\
\compactauthor{Dominic Buensalido}{2,†}
\compactauthor{Helen Kavnoudias}{2,†}
\compactauthor{Alain S. Abi-Ghanem}{3,†}
\compactauthor{Nour El Ghawi}{3,†}\\
\compactauthor{Cibele Luna}{4,†}
\compactauthor{Patricia Castillo}{5,†}
\compactauthor{Khaled Al-Surimi}{6,†}
\compactauthor{Rayyan A. Daghistani}{7,†}
\compactauthor{Yuh-Min Chen}{8,†}\\
\compactauthor{Heng-sheng Chao}{8,†}
\compactauthor{Lars Heiliger}{9,†}
\compactauthor{Moon Kim}{9,†}
\compactauthor{Johannes Haubold}{10,†}\\
\compactauthor{Frederic Jonske}{11,†}
\compactauthor{Pranav Rajpurkar}{1}
}
\end{minipage}

\vspace{1em}
\noindent\footnotesize{
\textsuperscript{*}Equal Contribution.\\
\textsuperscript{1}Department of Biomedical Informatics, Harvard Medical School, Boston, MA, USA\\
\textsuperscript{2}Department of Radiology, Alfred Health, Melbourne, Victoria, Australia\\
\textsuperscript{3}Department of Diagnostic Radiology, American University of Beirut, Beirut, Lebanon\\
\textsuperscript{4}Department of Radiology, University of Miami Miller School of Medicine, Miami, Florida, USA\\
\textsuperscript{5}University of Miami / Jackson Memorial Hospital, Miami, Florida, USA\\
\textsuperscript{6}Department of Healthcare Management, University of Doha for Science and Technology, Doha, Qatar\\
\textsuperscript{7}Department of Medical Imaging, King Abdulaziz Medical City, Riyadh, Saudi Arabia\\
\textsuperscript{8}Department of Chest Medicine, Taipei Veterans General Hospital, Taipei, Taiwan, Republic of China\\
\textsuperscript{9}Institute for AI in Medicine, University Hospital Essen, Essen, North Rhine-Westphalia, Germany\\
\textsuperscript{10}Department of Diagnostic and Interventional Radiology and Neuroradiology, University Hospital Essen, Essen, North Rhine-Westphalia, Germany\\
\textsuperscript{11}Department of Medical Machine Learning, Institute of AI in Medicine, University Medicine Essen, Essen, North Rhine-Westphalia, Germany\\
\textsuperscript{†}MAIDA Initiative Partners\\
[1em]
Oishi Banerjee: oishi\_banerjee@g.harvard.edu\\
Agustina Saenz: ads006@mail.harvard.edu\\
Kay Wu: kay.wu@medportal.ca\\
\\[1em]
}

\begin{abstract}
Given the rapidly expanding capabilities of generative AI models for radiology, there is a need for robust metrics that can accurately measure the quality of AI-generated radiology reports across diverse hospitals. We develop ReXamine-Global, a LLM-powered, multi-site framework that tests metrics across different writing styles and patient populations, exposing gaps in their generalization. First, our method tests whether a metric is undesirably sensitive to reporting style, providing different scores depending on whether AI-generated reports are stylistically similar to ground-truth reports or not. Second, our method measures whether a metric reliably agrees with experts, or whether metric and expert scores of AI-generated report quality diverge for some sites. Using 240 reports from 6 hospitals around the world, we apply ReXamine-Global to 7 established report evaluation metrics and uncover serious gaps in their generalizability. Developers can apply ReXamine-Global when designing new report evaluation metrics, ensuring their robustness across sites. Additionally, our analysis of existing metrics can guide users of those metrics towards evaluation procedures that work reliably at their sites of interest.

\end{abstract}


\section{Introduction}
\label{sec:intro}
The capabilities of AI are rapidly expanding in the field of radiology, with recent generative AI models comprehensively interpreting all aspects of radiology images and describing them in sophisticated text reports \cite{zhou2024, Hyland2023, Tanida2023, Tu2023}. To compare models and efficiently track progress in this space, developers rely heavily on automatic metrics that can efficiently score AI-generated radiology reports, measuring the accuracy of their content. These metrics measure the similarity between AI-generated candidate reports and ground-truth, radiologist-written reports; a candidate is assumed to be high-quality when metrics show it is similar to the corresponding ground-truth report. However, there are concerns that scores from commonly used metrics may not accurately evaluate the content of AI-generated reports, thus providing a misleading impression of model performance \cite{Yu2023}. Furthermore, automatic metrics have historically been used to evaluate models trained on and tested against reports from a handful of single-institution datasets \cite{Johnson2019, Demner-Fushman2016}, and it is unclear whether they generalize well across diverse reports from external sites.

\begin{figure}[htbp]
 % Caption and label go in the first argument and the figure contents
 % go in the second argument
  \centering
    \includegraphics[width=\linewidth]{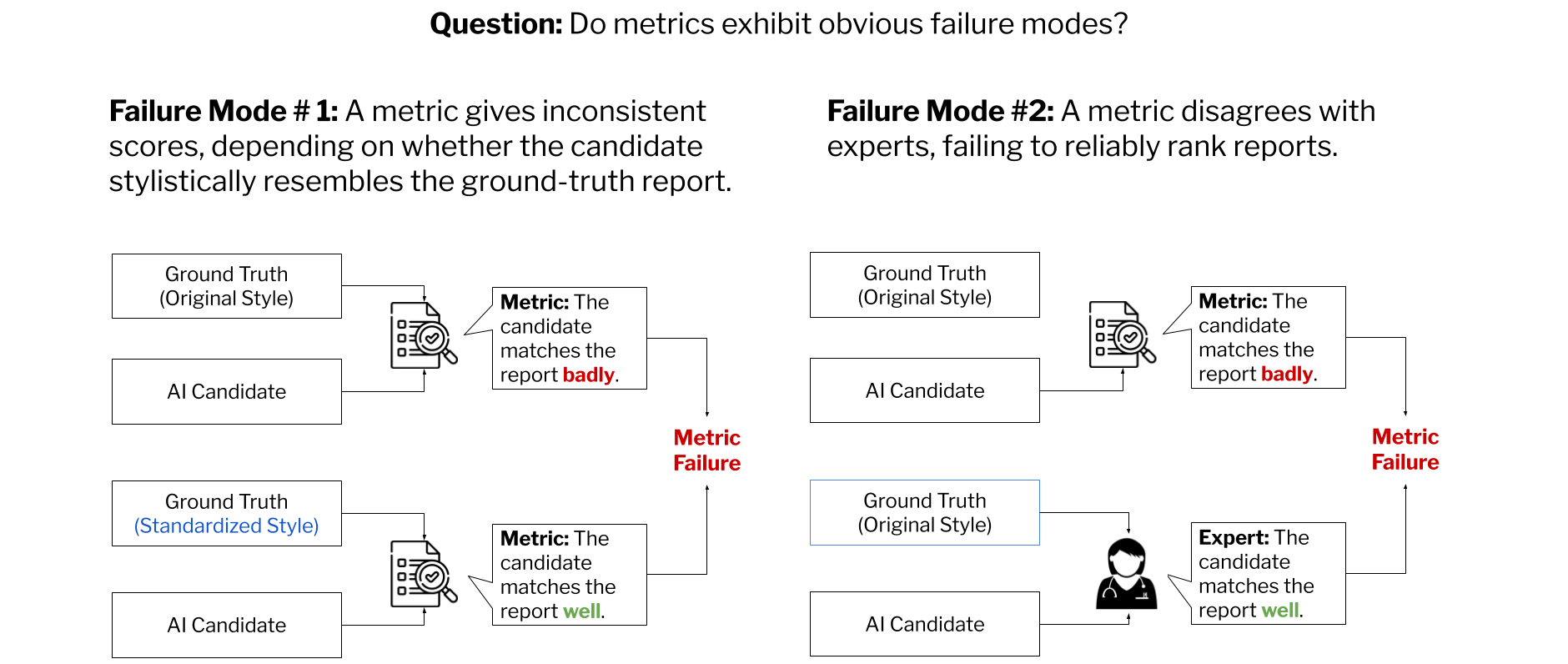}
    \caption{ReXamine-Global tests how metrics generalize when used across distributions, with the goal of uncovering two failure modes. First, we test whether automatic metrics are undesirably sensitive to clinically irrelevant differences in report style, providing different scores depending on whether candidates are stylistically similar to the ground truths. Next, we test whether metrics disagree with expert scores, providing unreliable judgments at some sites. A successful metric would avoid both failure modes.}
    \label{fig:experiments}
\end{figure}

In our work, we developed ReXamine-Global, a method for testing potential metrics across different writing styles and patient populations and exposing gaps in their generalizability. Using ground-truth reports from diverse hospitals, our method tests whether metrics are prone to two possible failure modes. First, we test whether metrics are undesirably sensitive to reporting style. Specifically, we explore whether they provide different scores depending on whether AI-generated reports are stylistically similar to ground-truth reports (e.g. during internal validation, when the model is tested against a familiar distribution) or not (as might occur during external validation, when model is tested against an unfamiliar distribution). Second, we check whether metric scores correlate with expert scores, with the expectation that an ideal metric would rank candidate reports exactly as an expert would. Using reports from 6 hospitals in different countries, we applied ReXamine-Global to test the generalizability of 7 established metrics for evaluating AI-generated radiology reports, revealing flaws in existing metrics.

Our work makes two primary contributions:
\begin{enumerate}
	\item We introduced ReXamine-Global, a new method for testing how report evaluation metrics generalize across diverse writing styles and patient populations. When creating new report evaluation metrics, developers can apply our method to determine whether metrics are overly sensitive to report-writing style or otherwise prone to poor generalization.
    \item By applying ReXamine-Global to 7 existing metrics, we uncovered gaps in the generalizability of many popular metrics, with a GPT-4-based metric outperforming all other approaches. These insights can help users of existing metrics design more reliable evaluation procedures for their sites of interest.
 
\end{enumerate}

\begin{table*}[htbp]
\centering
% \captionsetup{justification=raggedright, singlelinecheck=false}
\scriptsize	\setlength{\tabcolsep}{4pt}

\begin{tabular}{@{}l p{5.8cm} p{5.8cm}@{}}
\toprule
\textbf{Country} & \textbf{Example Reports} & \textbf{} \\
\midrule
Australia & 
ECMO catheter via inferior vena cava, tip in mid right atrium. Nasogastric tube in stomach. Left internal jugular central line tip in left brachiocephalic SVC junction. ETT 1 cm above carina. Left lower lobe collapse/consolidation. No pneumothorax or pleural effusion. &
ETT and pacemaker position. ETT tip 4 cm from carina. Increased density in left hemithorax consistent with pleural fluid collection. No consolidation seen. \\
\addlinespace
Germany & 
Rightly inserted endotracheal tube. Gastric tube subphrenically blanked out. Right transjugular CVC and sheath with tip projection to superior vena cava. New delineable sternal cerclages. Delineable clip material after mitral valve replacement. Progressive ateal confluent shading in left lung inferior field, mixed picture of pleural effusion and decreased ventilation. Increasing inferior ventilation in right lung subfield. Minor congestion signs. No pneumothorax. &
Heart and mediastinum widened in supine position. Patchy shadowing bipulmonary, likely due to congestion, concomitant atypical infiltrates cannot be excluded by projection radiography. Clinical correlation required. No major pleural effusion. No pneumothorax delineable in supine position. Properly inserted endotracheal tube. Transjugular CVC on right side with tip projection to superior vena cava. Gastric tube ending in projection onto left upper abdomen. \\
\addlinespace
Lebanon & 
Mild pulmonary edema. Cardiomegaly with cardiothoracic index of 0.57. No large pleural effusion or detectable pneumothorax. Single lead pacemaker with intact lead terminating in right ventricle topography. Chest wall intact. &
Increase in left basal pleural effusion with overlying haziness likely related to basal atelectasis. Right basal atelectatic bands. Right lung otherwise clear. No detectable right pleural effusion. Cardiac silhouette is in size. \\
\addlinespace
Saudi Arabia & 
Enlarged cardiac/pericardiac silhouette. Prominent central pulmonary vasculatures and bronchovascular markings suggest pulmonary congestion. Bilateral lower lung more of linear opacities may reflect atelectatic changes although infectious process not entirely excluded. &
Left upper lobe atelectatic band otherwise unremarkable study. \\
\addlinespace
Taiwan & 
Elevated right hemidiaphragm, tracheal deviated to Rt side. Right lung volume reduction is considered. Consolidation over right upper lung field, tumor growth cannot be r/o. R/o bullae over right lung apex &
Consolidation over right hemithorax, cause to be determined. Lung consolidation change and/or pleural effusion cannot be r/o. Trachea slightly deviated to Rt side. \\
\addlinespace
United States & 
IMPRESSION: Lines, tubes, etc: None. Cardiomediastinal silhouette: Within normal limits. Mediastinum midline. Lungs: Questionable subtle patchy right lower lung zone opacity which could represent an infectious process in the appropriate clinical setting, although limited due to overlying breast tissue summation. Pleura: Bilateral costophrenic angles sharp. No pneumothorax. Mild biapical pleural thickening/scarring. Bones/soft tissues: Unremarkable. &
IMPRESSION: Intact median sternotomy wires. Scattered surgical clips projecting over heart. Cardiac silhouette top normal in size. Trachea and mediastinum midline. Mild tortuosity of descending thoracic aorta. Greater than expected density of midline lower mediastinum, could reflect hiatal hernia, other lower mediastinal pathology not entirely excluded. No significant edema. No airspace consolidation. Mild asymmetric elevation of right hemidiaphragm. No appreciable pleural effusion or pneumothorax, though lung apex clipped from field-of-view. No aggressive osseous lesion. \\
\bottomrule
\end{tabular}
\small
\caption*{Table 1. Our dataset represents hospitals in 6 different countries, with reports that vary widely in content, terminology and organization. For example, the reports from Germany were automatically translated to English, resulting in atypical wording choices (e.g. ``delineable'', ``ateal"). Reports from Taiwan heavily featured abbreviations (e.g. ``Rt" for ``right"), while reports from the United States were longer than average, frequently containing several subsections. Variations such as these can pose a challenge for automatic metrics.}
\label{tab:medical-reports}

\end{table*}

\section{Methods}

\subsection*{The ReXamine-Global Framework} We proposed a LLM-powered framework for testing how a report evaluation metric performs across different writing styles and patient populations:
\begin{enumerate}
    \item \textbf{Multi-site data collection:} Gather a diverse dataset of ground-truth reports from multiple hospitals, representing a range of patient populations and writing styles.
    \item \textbf{Standardization of ground-truth texts:} Use a large language model (LLM) to rewrite the original ground-truth reports in a standardized style, while preserving the original content.
    \item \textbf{Generation of error-containing `candidate' texts:} Use a LLM to insert errors into standardized ground-truth reports. This step produces `candidate' reports, representing outputs from an imperfect radiology report generation model. 
    \item \textbf{Application of metric:} Use the metric to compare two pairs of reports: 1.) each candidate vs. its original ground-truth report (a stylistically different pair) and 2.) each candidate vs. its standardized ground-truth report (a stylistically similar pair).
    \item \textbf{Expert evaluation:} Engage clinical experts to manually evaluate the candidate reports, comparing them against ground-truth reports and counting the number of errors.
    \item \textbf{Assessment of metric consistency across styles:} Test whether, for any site, the metric produces significantly different scores for ``candidate-original" pairs and ``candidate-standardized" pairs. Ideally, a metric would always give a candidate the same score, regardless of whether it is being compared against the original or standardized ground-truth report.
    \item \textbf{Assessment of metric agreement with expert scores:} Test whether, for any site, the metric's scores fail to agree with expert scores. Ideally, metrics and experts will agree about which reports are the highest- and lowest-quality at every site, regardless of ground-truth style. 

\end{enumerate}
Using this framework, we assessed 7 existing automatic metrics for report evaluation.

\subsection*{Dataset} To apply ReXamine-Global, we  sampled reports from a private dataset containing chest X-ray reports from around the world, with a focus on emergency departments and intensive care units.  We selected reports that were either originally written in or translated into English. We included data from 6 hospitals in 6 different countries: United States, Saudi Arabia, Taiwan, Australia, Germany, and Lebanon. We randomly sampled 40 reports from each hospital, resulting in a total dataset of 240 reports. We refer to these reports as ``original ground-truth reports."

These radiology reports represent different patient populations as well as different writing styles, with marked differences in terminology, syntax and organization. For example, the reports from Germany were automatically translated to English, leaving artifacts that can prove challenging for automatic metrics. We give examples of these diverse reports in Table 1, which shows two examples from each site.

\subsection{Generation of Candidate Radiology Reports Using GPT-4}
After choosing 240 cases, we created 240 candidate reports, representing AI generations requiring evaluation. Our aim was to simulate outputs from an advanced but still flawed report generation model trained on MIMIC-CXR, a dataset widely used in the field \cite{johnson2019mimiccxrjpglargepubliclyavailable}. We used GPT-4 to produce a candidate report based on each radiologist-written ground-truth report, using a two-step process described further in Appendix A:
\begin{enumerate}
    \item \textbf{Standardizing Style:} Initially, GPT-4 was tasked with rewriting the `Findings' and `Impression' sections of a ground-truth report, using an example from MIMIC-CXR as a style guide. This step produced reports that preserved the original content but were written in a standardized, MIMIC-based style. A clinical expert checked 10 randomly sampled reports to ensure that this step did not change report content. We refer to these reports as ``standardized ground-truth reports."
    \item \textbf{Introducing Errors:} In the subsequent step, GPT-4 was instructed to deliberately introduce a few errors into the paraphrased report, thereby producing the final candidate report. We suggested several possible types of errors, such as the addition of a new finding, omission of an existing finding, or modification of the size or severity of a finding (Figure 2).
\end{enumerate}

\begin{figure}[htbp]
 % Caption and label go in the first argument and the figure contents
 % go in the second argument
  \centering
    \includegraphics[width=\linewidth]{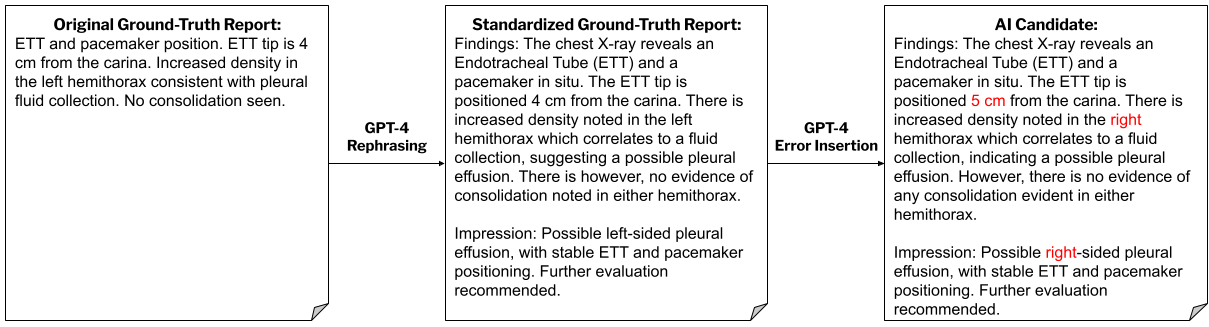}
    \caption{Using GPT-4, we first standardized the style of the ground-truth reports and then introduced errors to create AI candidates. For details on our prompts, please see Appendix A.}
    \label{fig:prompt}
\end{figure}

\subsection{Automatic Metrics}
We examined seven existing automatic metrics used to judge the quality of AI-generated radiology reports. We included two general-purpose metrics that are not specialized for medical text: BLEU-2, which counts overlapping substrings in the ground-truth text and AI-generated text \cite{Papineni2002}, and BERTScore, which computes the similarity of embeddings produced by passing each text through a general-purpose BERT model \cite{Zhang2019} . Additionally, we considered clinical metrics such as CheXbert vector similarity, which compares the similarity of embeddings produced by passing each text through a specialized medical BERT model \cite{smit2020chexbertcombiningautomaticlabelers}, and RadGraph-F1, which uses a specialized medical model to extract a graph of medical entities and relations from each text and measures the similarity of the graphs \cite{jain2021radgraphextractingclinicalentities}. Additionally, we studied two versions of the RadCliQ metric, recently proposed specifically for evaluating AI-generated reports \cite{Yu2023}. RadCliQ-v0 and RadCliQ-v1 both use a machine learning model to take in values from other metrics, such as BERTScore and CheXbert vector similarity, and then produce a composite score based on these input values. Finally, we considered FineRadScore, a recently proposed method that uses LLMs to perform a line-by-line comparison of ground-truth and candidate reports \cite{huang2024fineradscore}. In our implementation of FineRadScore, we used GPT-4 to identify lines requiring corrections and treated the total number of problematic lines as the final score, which we refer to FineRadScore-GPT-4.

\subsection{Expert Evaluation}
To obtain gold-standard measurements of candidate report quality, we conducted a manual evaluation engaging both an internal medicine attending and a radiology resident. The evaluation protocol was based on a scoring system adapted from the American College of Radiology \cite{Goldberg-Stein2017} and from prior research studies \cite{Yu2023}, designed to assess the clinical significance of discrepancies in report interpretations. Errors were classified into seven independent categories: False prediction of finding; Omission of finding;  Incorrect location of finding; Incorrect position of finding; Incorrect severity of finding. Mention of comparison that is not present in the reference impression; Omission of comparison describing a change from a previous study. We counted the total number of errors found in each report to produce our final expert score, so lower-quality candidates receive higher scores. For this study, each reviewer was assigned 120 unique reports, with an additional 10 reports each to assess inter-rater agreement.

\subsection{Experiments}

We used our 7 automatic metrics and expert evaluation to compare two types of report pairs: (1) the original ground-truth report vs. the AI candidate report, and (2) the standardized ground-truth report vs. the AI candidate. We assessed how automatic metrics performed on these comparisons using two approaches. First, we tested whether AI candidates received different scores when compared against the standardized ground-truth report rather than the original ground-truth report; we assume an ideal metric would be robust against clinically irrelevant stylistic variations and therefore give the same scores in both experiments. Second, we tested whether metric scores agreed with expert scores, as an ideal metric would provide the same ranking of a site's reports as experts do. These two approaches allowed us to compare how metrics behave when assessing reports with different styles (original ground truth vs. AI candidate) and reports with similar styles (standardized ground truth vs. AI candidate), as the standardized ground truth and AI candidate reports share a common GPT-4-generated style. 

To facilitate interpretation of our results, we standardized the directionality of all automatic and human evaluation metrics, so that a higher score consistently indicates worse performance from the report generation model. Originally, higher scores for BLEU-2, BERTScore, CheXbert vector similarity, and RadGraph-F1 indicated better performance, while lower scores for RadCliQ and FineRadScore-GPT-4 indicated better performance. To align all metrics so a higher score indicates worse performance, we multiplied the scores of BLEU-2, BERTScore, CheXbert vector similarity, and RadGraph-F1 by -1. This standardization makes it easier to compare our results across different evaluation metrics.

We employed two main statistical approaches to study the behavior of automatic metrics across different countries and ground-truth styles. 
First, we conducted paired t-tests to determine whether automatic metrics provide different scores depending on whether original or standardized ground-truth reports are used. These tests were performed independently for each country to account for potential regional variations. To address the issue of multiple comparisons in our t-test analyses, we applied a Bonferroni correction to control the familywise error rate. The significance level $\alpha$ was set at 0.05, and the Bonferroni-corrected threshold was calculated as $\alpha/n$, where $n$  is the total number of paired t-tests conducted (number of metrics × number of countries = 42).
Second, we calculated Spearman's rank correlation coefficients ($\rho$) to quantify the agreement between automatic metrics and human evaluations for each country. This analysis was performed separately when using original and standardized ground-truth reports, allowing us to assess how well our automatic metrics aligned with human judgments across different ground-truth styles and geographical regions.

\section{Results}

\subsection{Effect of Stylistic Differences on Metric Scores}

We found that stylistic differences significantly impacted scores from all metrics, with the exception of FineRadScore-GPT-4. Across all non-GPT metrics and countries, paired t-tests revealed significant differences in scores depending on whether original or standardized ground-truth reports were used (Bonferroni-corrected $p < 0.05$) (Table 2). BERTScore showed the highest mean t-statistics across all countries (mean t-stat = -29.72, range: -17.24 to -37.09), indicating a substantial and consistent difference in scores between the two report styles. FineRadScore-GPT-4 exhibited the smallest t-statistics (mean t-stat = -1.07, range: -1.50 to -0.42) and was the only metric that did not show significant differences for any country after Bonferroni correction.
All t-statistics were negative, indicating that comparisons between standardized ground truth reports and AI candidates consistently yielded lower scores (i.e. indicating higher-quality AI candidates) compared to comparisons between original ground-truth reports and AI candidates. In other words, metrics rated the AI model as better-performing when the ground truth stylistically resembled the AI candidate.

\begin{table}[h]
\centering
\caption{Summary of paired t-test results across metrics}
\label{tab:t_test_summary}
\begin{tabular}{lrrrr}
\toprule
Metric & Mean t-stat & Min t-stat & Max t-stat & Significant Countries \\
\midrule
BLEU-2 \cite{Papineni2002} & -27.23 & -31.01 & -20.60 & 6 \\
BERTScore \cite{Zhang2019} & -29.72 & -37.09 & -17.24 & 6 \\
CheXbert Similarity \cite{smit2020chexbertcombiningautomaticlabelers} & -6.29 & -8.15 & -3.97 & 6 \\
RadCliQ-v0 \cite{Yu2023}& -20.50 & -30.08 & -11.20 & 6 \\
RadCliQ-v1 \cite{Yu2023}& -22.23 & -32.37 & -12.77 & 6 \\
RadGraph-F1 \cite{jain2021radgraphextractingclinicalentities} & -13.66 & -19.18 & -9.65 & 6 \\
\textbf{FineRadScore-GPT-4 \cite{huang2024fineradscore}} & \textbf{-1.07} & \textbf{-1.50} & \textbf{-0.42} & \textbf{0} \\
\bottomrule
\end{tabular}
\caption*{Table 2. Negative t-statistics indicate that standardized ground truth-AI candidate pairs (similar styles) consistently received lower scores than original ground truth-AI candidate pairs (different styles). The magnitude of the t-statistic reflects the strength of this difference. The ``Mean'' value gives the average t-statistic across all 6 countries, while he ``Min" and ``Max" t-stat values show the lowest and highest values seen across the 6 countries. The ``Significant Countries" column indicates the number of countries (out of 6) where the metric showed a significant difference between ground truth-AI candidate and standardized ground truth-AI candidate pairs after Bonferroni correction. FineRadScore-GPT-4 is the only metric whose scores were not significantly affected by the ground-truth style.}
\end{table}

\begin{figure}[ht]
\centering
\includegraphics[width=.9\textwidth]{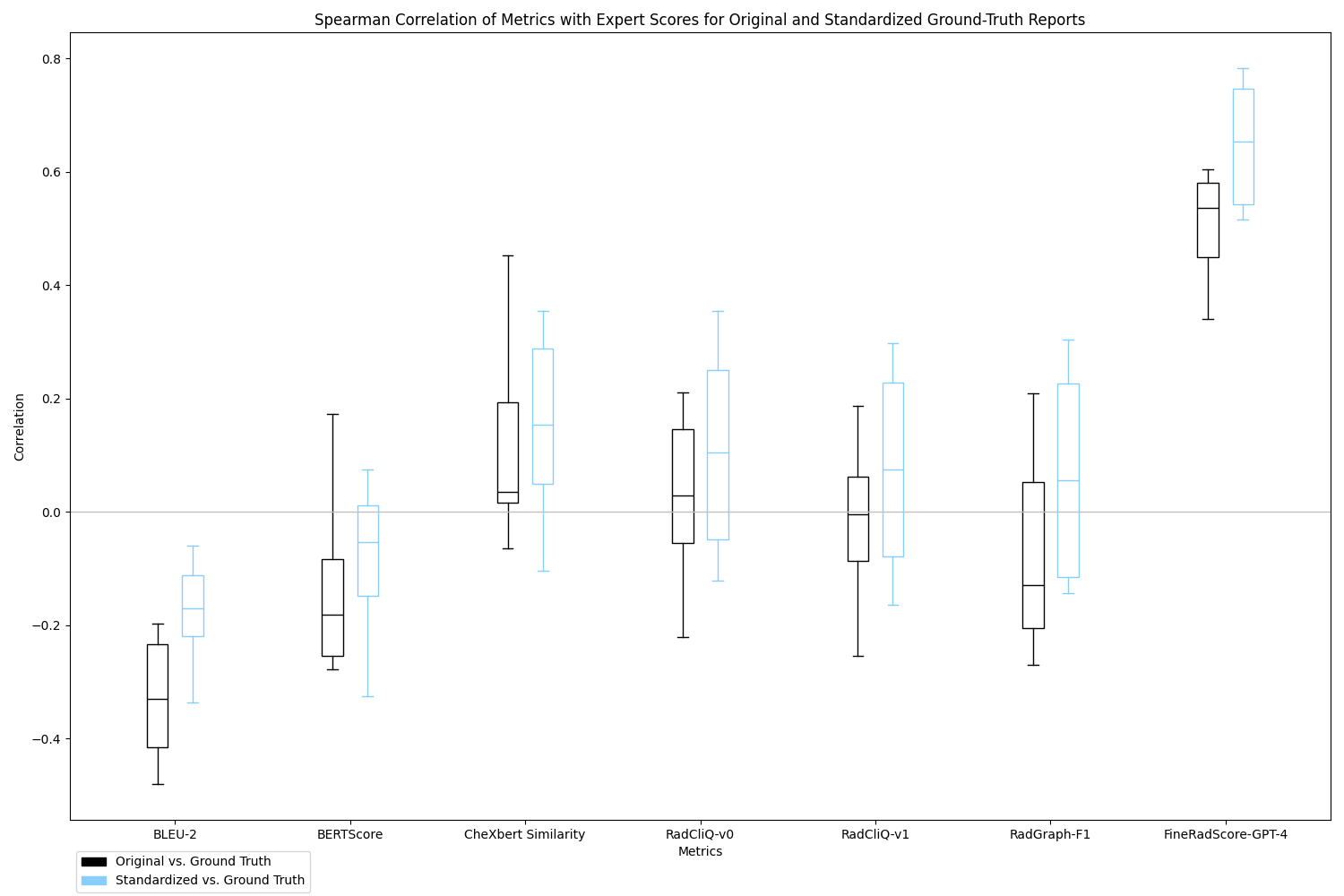}
\caption{Except for FineRadScore-GPT-4, no metric achieved positive Spearman correlations with expert scores at every site, indicating poor generalization. Correlations for original ground-truth reports are shown in the black box plots (left). Correlations for standardized ground-truth reports are shown in blue box plots (right). Metrics typically achieved higher performance with standardized ground-truth reports than original ground-truth reports. For detailed numerical results, see the table in Appendix C.}

\label{fig:t_stat_significance_plot}
\end{figure}

\subsection{Correlation of Automatic Metrics with Expert Scores on Stylistically Diverse Reports}
When comparing original ground-truth reports against stylistically different candidates, metrics frequently failed to align with experts (Figure 3). FineRadScore-GPT-4, the only metric using a LLM, offered the best performance, with coefficients ranging from ($\rho = 0.34$ to $0.60$). Despite achieving positive correlations at some sites, each of the other metrics had negative coefficients for at least one site. BLEU-2 showed especially poor performance, with Spearman's rank correlation coefficients ($\rho$) ranging from $\rho = -0.20$ to $-0.48$. Information on expert scores can be found in Appendix B.

\subsection{Correlation of Automatic Metrics with Expert Scores on Stylistically Standardized Reports}
After standardizing ground-truth reports to resemble the style of the candidates, metrics generally showed better agreement with experts (Figure 3). For example, FineRadScore-GPT-4's coefficients rose across all sites, now ranging from $\rho = 0.52$ to $0.78$. Despite similar increases, every other metric still had a negative coefficient for at least one site, suggesting that metrics can fail to generalize even after standardization. Notably, BLEU-2's correlation coefficients remained consistently negative even after standardization, ranging from $\rho = -0.34$ to $-0.06$.

\section{Discussion}

ReXamine-Global, which tests report evaluation metrics across diverse distributions, successfully revealed critical gaps in metric generalizability. By applying ReXamine-Global to 7 existing metrics, we found that most automatic metrics are undesirably sensitive to stylistic differences, giving significantly different scores depending on the style of the ground-truth report. The only exception was FineRadScore-GPT-4, which used a powerful LLM to evaluate reports \cite{huang2024fineradscore}. Furthermore, we observed that automatic metrics of all kinds demonstrated, at best, moderate correlation with expert opinions when using original ground-truth reports. Metrics generally attained better correlations when comparing candidates against standardized ground-truth reports, opening the possibility that preprocessing candidates and ground-truth reports to make them stylistically similar can improve evaluation procedures. Importantly, we observed that metric behavior sometimes varied across hospitals; for example, CheXbert Similarity's correlations when comparing candidates and original ground-truth reports ranged from -0.065 to 0.45. This finding shows the importance of including data from a range of diverse hospitals.

The clear variability in metric performance across sites highlights important directions for future work. ReXamine-Global can guide the development of more robust report evaluation metrics, capable of generalizing effectively across diverse healthcare settings. Additionally, we hope our work can warn users about the risks of naively applying metrics to new distributions and help them choose high-performing metrics for their specific sites of interest.

\subsection{Limitations}
While we utilized GPT-4 to generate standardized ground-truth and candidate reports, candidate reports generated by other models may elicit different behavior from metrics, so a metric that performs well on ReXamine-Global may generalize poorly to some other distribution. In addition, our manual evaluation scoring system did not encompass all possible error categories, potentially overlooking some types of inaccuracies, and our evaluation was conducted by only two physicians, which significantly limits the breadth and diversity of expert assessment. This constraint may have introduced bias and reduced the robustness of our manual evaluation results. Ideally, each report would be reviewed by multiple physicians from diverse specialties, with a third reviewer to resolve discrepancies. This approach would provide a more comprehensive and reliable assessment of report quality and error identification. A larger pool of reviewers would also make it possible to conduct inter-rater reliability analyses, which could confirm the consistency and reliability of manual evaluation.

\section{Institutional Review Board (IRB)}\label{sec:irb}
All data was obtained following the approval of Institutional Review Board (IRB) and Data Use Agreement (DUA) protocols. 
\section{Data Contribution}
The following members of the MAIDA Initiative made substantial contributions to data collection for this study:
Dominic Buensalido, Warren Clements, Helen Kavnoudias, Adil Zia, Nour El Ghawi, Alain S. Abi-Ghanem, Khalid Al-Surimi, Rayyan A. Daghistani, Moon Kim, Frederic Jonske, Johannes Haubold, Lars Heiliger, Rosa Patricia Castillo, Cibele C. Luna, Heng-sheng Chao, and Yuh-Min Chen. More information about the MAIDA initiative can be found at https://www.rajpurkarlab.hms.harvard.edu/maida.

\section{Acknowledgments}
We thank Xiaoman Zhang for figure design assistance, Wendy Erselius for managing data collection and Jonathan Dreyfuss for statistical analysis support. This work was supported by Biswas Family Foundation's Transformative Computational Biology Grant in Collaboration with the Milken Institute.
\newpage

\section{Appendices}

\appendix{GPT-4 Instructions}

We gave GPT-4 the following instructions when standardizing the style of our original ground-truth reports:

\begin{verbatim}
    Pretend you are a radiologist and format the content of these notes in a 
    polished findings and impressions section. Your findings section may be 
    long or short. Your impression should only have 1-3 lines. If you are 
    unsure about an abbreviation, term, or other odd phrasing, make your best 
    guess. Match the style of this radiology report:

    Report:
    Findings: Single frontal view of the chest demonstrates a right 
    Port-A-Cath in unchanged position, terminating at the cavoatrial junction.
    Median sternotomy wires are present, along with surgical clips in the left 
    upper quadrant.  The heart is mildly enlarged, but stable compared with 
    prior examinations, with redemonstration of calcified mediastinal lymph
    nodes. A rounded opacity in the lower left lung likely correlates to a 
    calcified granuloma as seen on CT of the chest from ___.  There is no 
    evidence of pneumonia, pleural effusion, pneumothorax or overt pulmonary 
    edema.  The lung volumes are low, accentuating bibasilar atelectasis.  No
    subdiaphragmatic free air is present.	

    Impression: No subdiaphragmatic free air or other acute cardiopulmonary 
    process.
\end{verbatim}

After standardizing the style of our reports, we used the following instructions to introduce errors, producing the final candidate:
\begin{verbatim}
    Please write a report using the above report as a template. Perturb the 
    content of a few existing lines. Here are some examples of how a line 
    could be changed:
    - If the report says X condition is present, state that X condition is 
    absent.
    - If the report rules out X condition, state that X condition is present. 
    - Change the location, size, severity, or implications of a condition.
 
    Only perturb a few lines. Keep the other lines exactly the same. Your 
    report should still sound fluent, like a radiologist wrote it.
\end{verbatim}

\newpage

\appendix{Manual Evaluation Results}

This table gives more details of our manual evaluation, providing the number of errors identified across the 40 candidate reports generated each of the different hospital sites.

\begin{table}[hbtp]
\centering
  {\caption{Total number of all errors and clinically significant errors made across candidate reports for each hospital site, as per reviewer count.}}
  {\begin{tabular}{>{\raggedright\arraybackslash}p{0.3\linewidth} >{\raggedright\arraybackslash}p{0.3\linewidth} >{\raggedright\arraybackslash}p{0.3\linewidth}}
  \toprule
  \bfseries Hospital Site & \bfseries Total number of errors \\
  \midrule

  US & 138 \\
  Germany & 100 \\
  Lebanon & 103 \\
  Taiwan & 126 \\
  Australia & 106 \\
  Saudi Arabia & 94 \\
  \bottomrule
  \end{tabular}}
\end{table}

\appendix{Full Correlation Results}

This table gives detailed results about how metric scores were correlated with expert scores, across sites and ground-truth report styles.

\begin{table}[htbp]
    \centering
    \scriptsize
    \caption{A Table with 7 Columns and 14 Rows}
    \begin{tabular}{|p{2.9cm}|p{2cm}|p{1.2cm}|p{1.2cm}|p{1.2cm}|p{1.2cm}|p{1.2cm}|p{1.2cm}|}
        \hline
        
        \textbf{Metric} & \textbf{Ground Truth} & \textbf{Australia} &\textbf{Lebanon} &\textbf{Taiwan} &\textbf{Saudi Arabia} &\textbf{United States} &\textbf{Germany}  \\
        \hline
        \multirow{2}{*}{\textbf{BLEU-2}}  & Original & -0.48 & -0.44 & -0.35 & -0.20 & -0.31 & -0.21 \\
         & Standardized & -0.10 & -0.34 & -0.06 & -0.20 & -0.23 & -0.15 \\
        \hline
        \multirow{2}{*}{\textbf{BERTScore}} & Original & -0.26 & -0.28 & -0.07 & -0.25 & -0.11 & 0.17 \\
         & Standardized & 0.07 & -0.33 & -0.02 & 0.02 & -0.17 & -0.08 \\
        \hline
        \multirow{2}{*}{\textbf{CheXbert Similarity}} & Original & 0.24 & -0.06 & 0.45 & 0.03 & 0.01 & 0.04 \\
         & Standardized & 0.36 & -0.10 & 0.30 & 0.24 & 0.06 & 0.04 \\
        \hline
        \multirow{2}{*}{\textbf{RadCliQ-v0}} & Original & 0.06 & -0.22 & 0.17 & -0.00 & -0.07 & 0.21 \\
          & Standardized & 0.35 & -0.12 & 0.25 & 0.25 & -0.05 & -0.04 \\
        \hline
        \multirow{2}{*}{\textbf{RadCliQ-v1}} & Original & 0.00 & -0.25 & 0.08 & -0.01 & -0.11 & 0.19 \\
          & Standardized & 0.30 & -0.16 & 0.24 & 0.20 & -0.09 & -0.05 \\
        \hline
        \multirow{2}{*}{\textbf{RadGraph-F1}} & Original & -0.06 & -0.27 & -0.21 & 0.09 & -0.20 & 0.21 \\
          & Standardized & 0.17 & -0.14 & 0.30 & 0.24 & -0.13 & -0.06 \\
        \hline
        \multirow{2}{*}{\textbf{FineRadScore-GPT-4}} & Original & 0.56 & 0.59 & 0.51 & 0.60 & 0.43 & 0.34 \\
         & Standardized & 0.78 & 0.76 & 0.62 & 0.52 & 0.52 & 0.69 \\
        \hline
    \end{tabular}
\end{table}
\newpage

\nocite{*} % to test all bib entrys
\bibliographystyle{unsrt}
\bibliography{main} 

\begin{thebibliography}{10}

\bibitem{zhou2024}
Hong-Yu Zhou, Subathra Adithan, Julián~Nicolás Acosta, Eric~J. Topol, and Pranav Rajpurkar.
\newblock A generalist learner for multifaceted medical image interpretation, 2024.

\bibitem{Hyland2023}
Stephanie~L Hyland, Shruthi Bannur, Kenza Bouzid, Daniel~C Castro, Mercy Ranjit, Anton Schwaighofer, Fernando P{'e}rez-Garc{'i}a, et~al.
\newblock Maira-1: A specialised large multimodal model for radiology report generation.
\newblock {\em arXiv preprint arXiv:2311.13668}, 2023.

\bibitem{Tanida2023}
Tim Tanida, Philip M{"u}ller, Georgios Kaissis, and Daniel Rueckert.
\newblock Interactive and explainable region-guided radiology report generation.
\newblock In {\em Proceedings of the IEEE/CVF Conference on Computer Vision and Pattern Recognition (CVPR)}, 2023.

\bibitem{Tu2023}
Tao Tu, Shekoofeh Azizi, Danny Driess, Mike Schaekermann, Mohamed Amin, Pi-Chuan Chang, Andrew Carroll, et~al.
\newblock Towards generalist biomedical ai.
\newblock {\em arXiv preprint arXiv:2307.14334}, 2023.

\bibitem{Yu2023}
Feiyang Yu, Mark Endo, Rayan Krishnan, Ian Pan, Andy Tsai, Eduardo~Pontes Reis, Eduardo Kaiser~Ururahy Fonseca, et~al.
\newblock Evaluating progress in automatic chest x-ray radiology report generation.
\newblock {\em Patterns}, 2023.

\bibitem{Johnson2019}
Alistair E~W Johnson, Tom~J Pollard, Seth~J Berkowitz, Nathaniel~R Greenbaum, Matthew~P Lungren, Chih-Ying Deng, Roger~G Mark, and Steven Horng.
\newblock Mimic-cxr, a de-identified publicly available database of chest radiographs with free-text reports.
\newblock {\em Scientific Data}, 6(1):317, 2019.

\bibitem{Demner-Fushman2016}
Dina Demner-Fushman, Marc~D Kohli, Marc~B Rosenman, Sonya~E Shooshan, Laritza Rodriguez, Sameer Antani, George~R Thoma, and Clement~J McDonald.
\newblock Preparing a collection of radiology examinations for distribution and retrieval.
\newblock {\em Journal of the American Medical Informatics Association: JAMIA}, 23(2):304--310, 2016.

\bibitem{johnson2019mimiccxrjpglargepubliclyavailable}
Alistair E.~W. Johnson, Tom~J. Pollard, Nathaniel~R. Greenbaum, Matthew~P. Lungren, Chih ying Deng, Yifan Peng, Zhiyong Lu, Roger~G. Mark, Seth~J. Berkowitz, and Steven Horng.
\newblock Mimic-cxr-jpg, a large publicly available database of labeled chest radiographs, 2019.

\bibitem{Papineni2002}
Kishore Papineni, Salim Roukos, Todd Ward, and Wei-Jing Zhu.
\newblock Bleu: A method for automatic evaluation of machine translation.
\newblock In {\em Annual Meeting of the Association for Computational Linguistics}, pages 311--318, 2002.

\bibitem{Zhang2019}
Tianyi Zhang, Varsha Kishore, Felix Wu, Kilian~Q Weinberger, and Yoav Artzi.
\newblock Bertscore: Evaluating text generation with bert.
\newblock {\em arXiv preprint arXiv:1904.09675}, 2019.

\bibitem{smit2020chexbertcombiningautomaticlabelers}
Akshay Smit, Saahil Jain, Pranav Rajpurkar, Anuj Pareek, Andrew~Y. Ng, and Matthew~P. Lungren.
\newblock Chexbert: Combining automatic labelers and expert annotations for accurate radiology report labeling using bert, 2020.

\bibitem{jain2021radgraphextractingclinicalentities}
Saahil Jain, Ashwin Agrawal, Adriel Saporta, Steven~QH Truong, Du~Nguyen Duong, Tan Bui, Pierre Chambon, Yuhao Zhang, Matthew~P. Lungren, Andrew~Y. Ng, Curtis~P. Langlotz, and Pranav Rajpurkar.
\newblock Radgraph: Extracting clinical entities and relations from radiology reports, 2021.

\bibitem{huang2024fineradscore}
Alyssa Huang, Oishi Banerjee, Kay Wu, Eduardo~Pontes Reis, and Pranav Rajpurkar.
\newblock Fineradscore: A radiology report line-by-line evaluation technique generating corrections with severity scores, 2024.

\bibitem{Goldberg-Stein2017}
Shlomit Goldberg-Stein, L~Alexandre Frigini, Scott Long, Zeyad Metwalli, Xuan~V Nguyen, Mark Parker, and Hani Abujudeh.
\newblock Acr radpeer committee white paper with 2016 updates: Revised scoring system, new classifications, self-review, and subspecialized reports.
\newblock {\em Journal of the American College of Radiology: JACR}, 14(8):1080--1086, 2017.

\end{thebibliography}

\end{document}